\documentclass[11pt, a4paper,  notitlepage]{extarticle}

\usepackage[left=3.5cm, right=2.5cm, top=3cm, bottom=3cm, bindingoffset=0cm]{geometry}
\usepackage{graphicx}
\usepackage{mathtools}
\usepackage[affil-it]{authblk}
\usepackage{url}
\usepackage{amsmath}
\usepackage{amssymb}
\usepackage{multicol}
\usepackage{float}
\usepackage[section]{placeins}
\usepackage{hyperref}
\usepackage[title]{appendix}
\usepackage{changepage}
\usepackage{listings}

\begin{document}
\sloppy
\title{Predictive Analytics of Air Alerts in the Russian-Ukrainian War}
\author{Demian Pavlyshenko, Bohdan Pavlyshenko \\  \small{demkolviv@gmail.com, b.pavlyshenko@gmail.com}}
\maketitle

\begin{abstract}
The paper considers exploratory data analysis and approaches in predictive analytics for air alerts during the Russian-Ukrainian war which broke out on Feb 24, 2022.  The results illustrate that alerts in regions correlate with one another and have geospatial patterns which
make it feasible to build a predictive model which predicts alerts that are expected to take place in a certain region within a specified time period.
The obtained results show that the alert status in a particular region is highly dependable on the features of its adjacent regions.
Seasonality features like hours, days of a week and months are also crucial in predicting the target variable. Some regions highly rely on the time feature which equals to a number of days from the initial date of the dataset. From this, we can deduce that the air alert pattern changes throughout the time.

\end{abstract}


\section{Introduction}
Starting from February 24, 2022, the date of Russian invasion in Ukraine, it has been very important to understand the structure and patterns of air alerts and predict when and how long an air alert is going to take place. Initially, we created a channel in  Telegram Messenger social network for air alerts forecast in Ukraine. The main approach was based on loading messages from  other Telegram Messenger channels, analyze them using NLP methods, and then, using experimental heuristics, make prediction when  an air alert is about to start.
	Currently, there are many similar channels in Telegram Messenger, which publish up-to-date information about current air alerts. At the same time, there are many datasets with historical data about air alerts. Our experience of air alerts intuitively shows that  there is a geospatial pattern in emerging alerts in different regions of Ukraine. As a result, knowing the cause of the alert and how alerts propagated in different regions, we can anticipate when an air alert is going to start and how long it is going to last in our region. Air alerts analytics is also considered in~\cite{airraiddatasets, van2023public, airalarmsinua, appairalert, airraidmonitor,  vizairalertkaggle, vizairalert, buildingairalarm}.
	
	The main goal of our study is to conduct an exploratory data analysis and create a predictive model to forecast the duration of air alerts.
\section{Exploratory data analysis}
\begin{figure}[H]
\center
 \includegraphics[width=0.75\linewidth]{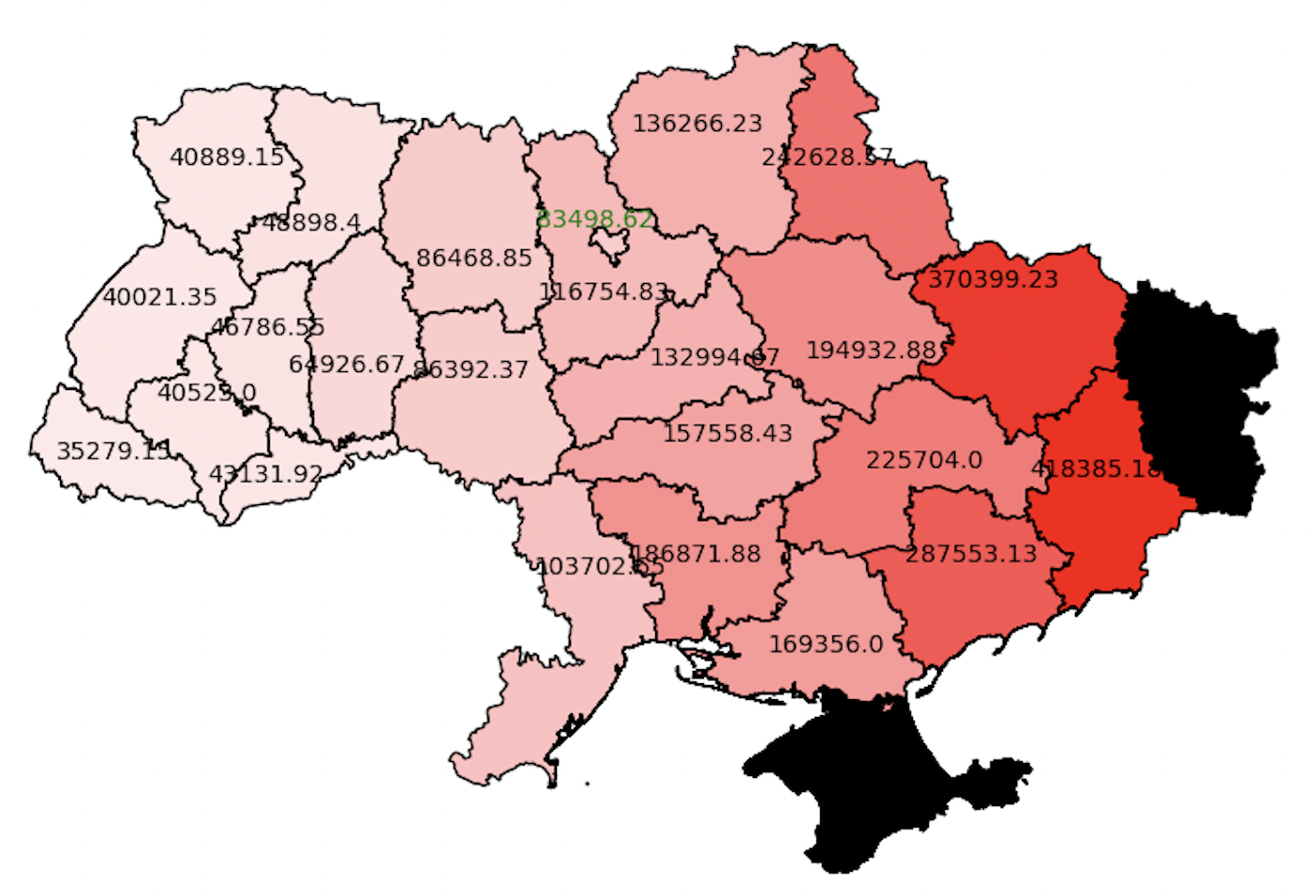}
 \caption{Heatmap for the total duration of air alerts in Ukrainian regions (minutes)}
 \label{time_distribution}
 \end{figure}

\begin{figure}[H]
\center
 \includegraphics[width=0.75\linewidth]{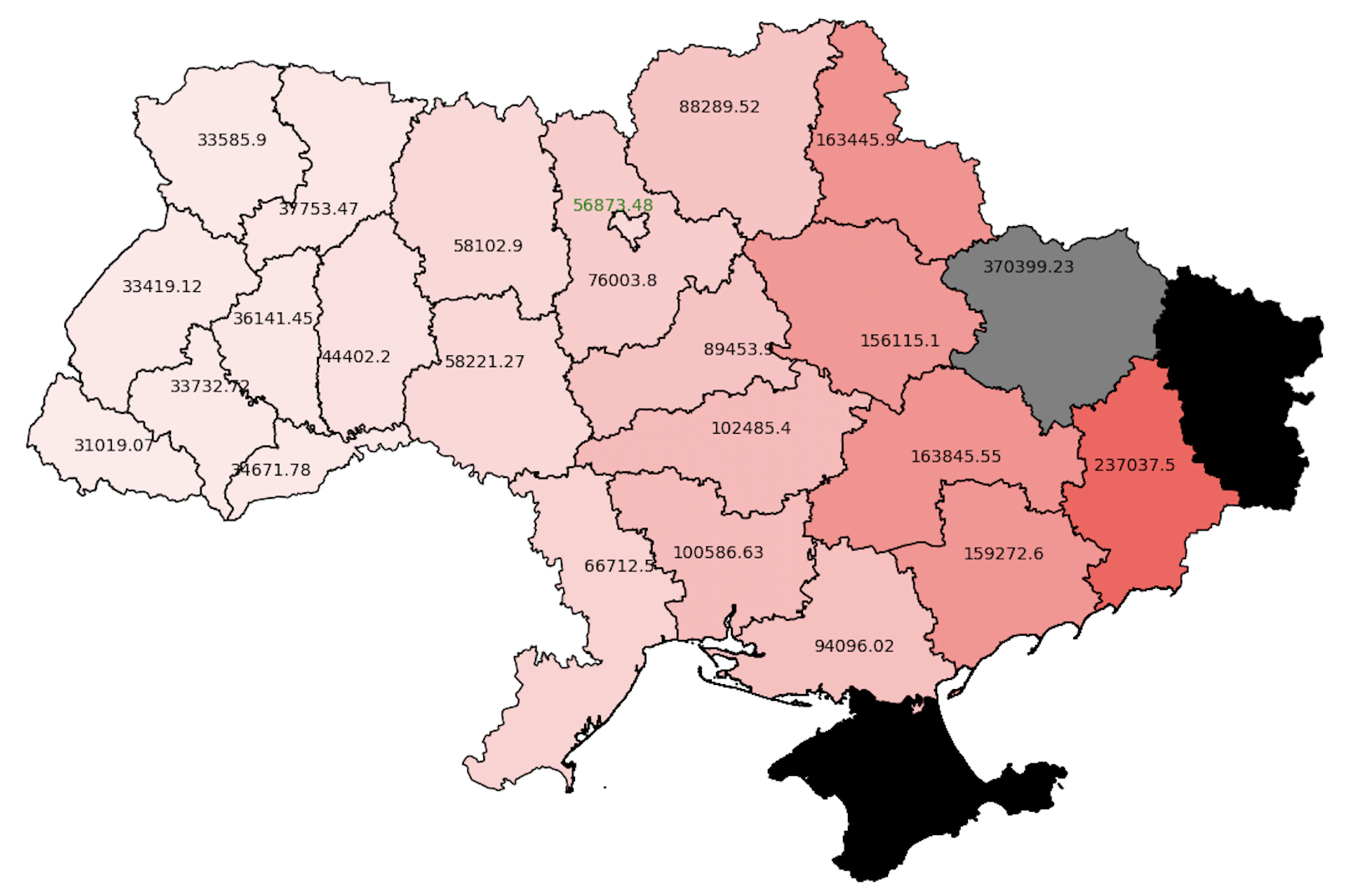}
 \caption{Heatmap for regions' total duration of air alerts which occurred simultaneously with Kharkiv oblast (minutes)}
 \label{kharkiv_alert_conjc}
 \end{figure}

\begin{figure}[H]
\center
 \includegraphics[width=0.85\linewidth]{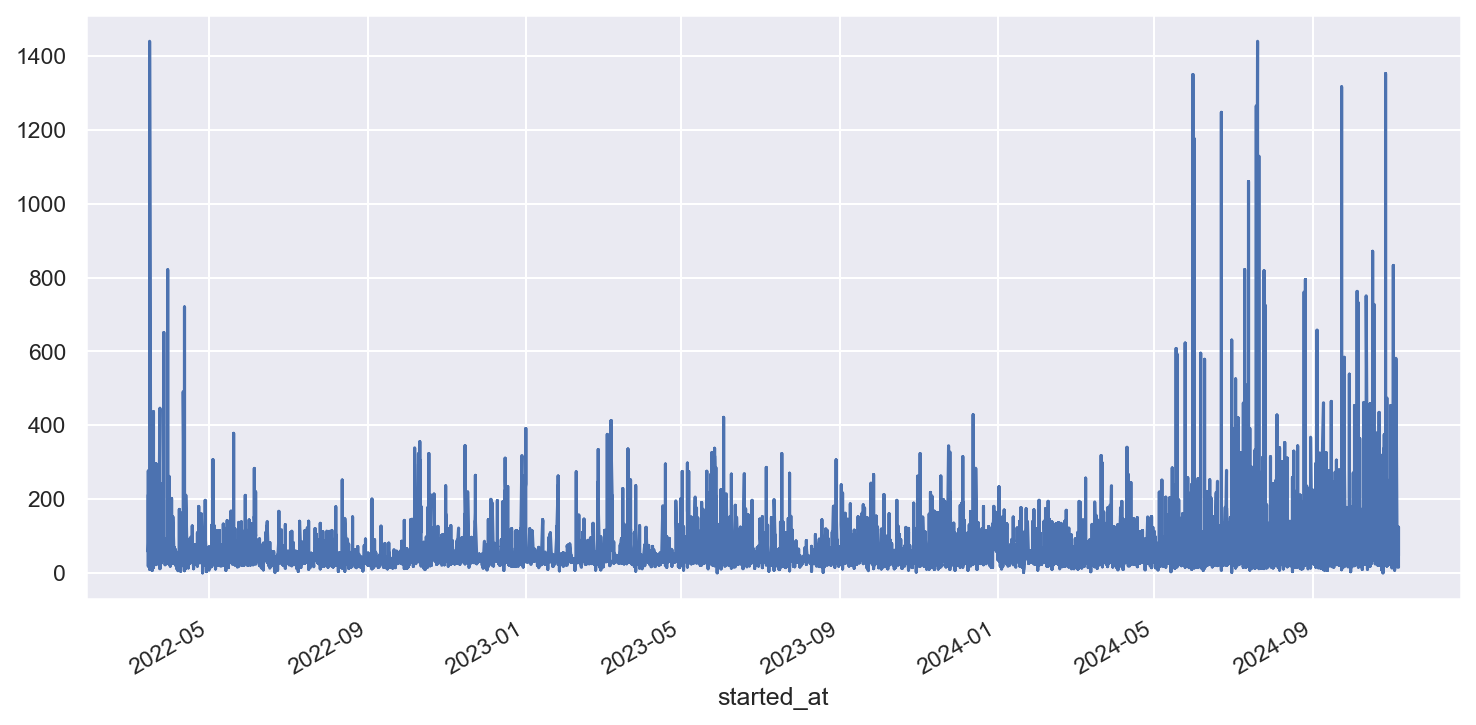}
 \caption{Time series for daily median for alert durations by regions (minutes)}
 \label{median_time_diff}
 \end{figure}

\begin{figure}[H]
\center
 \includegraphics[width=0.7\linewidth]{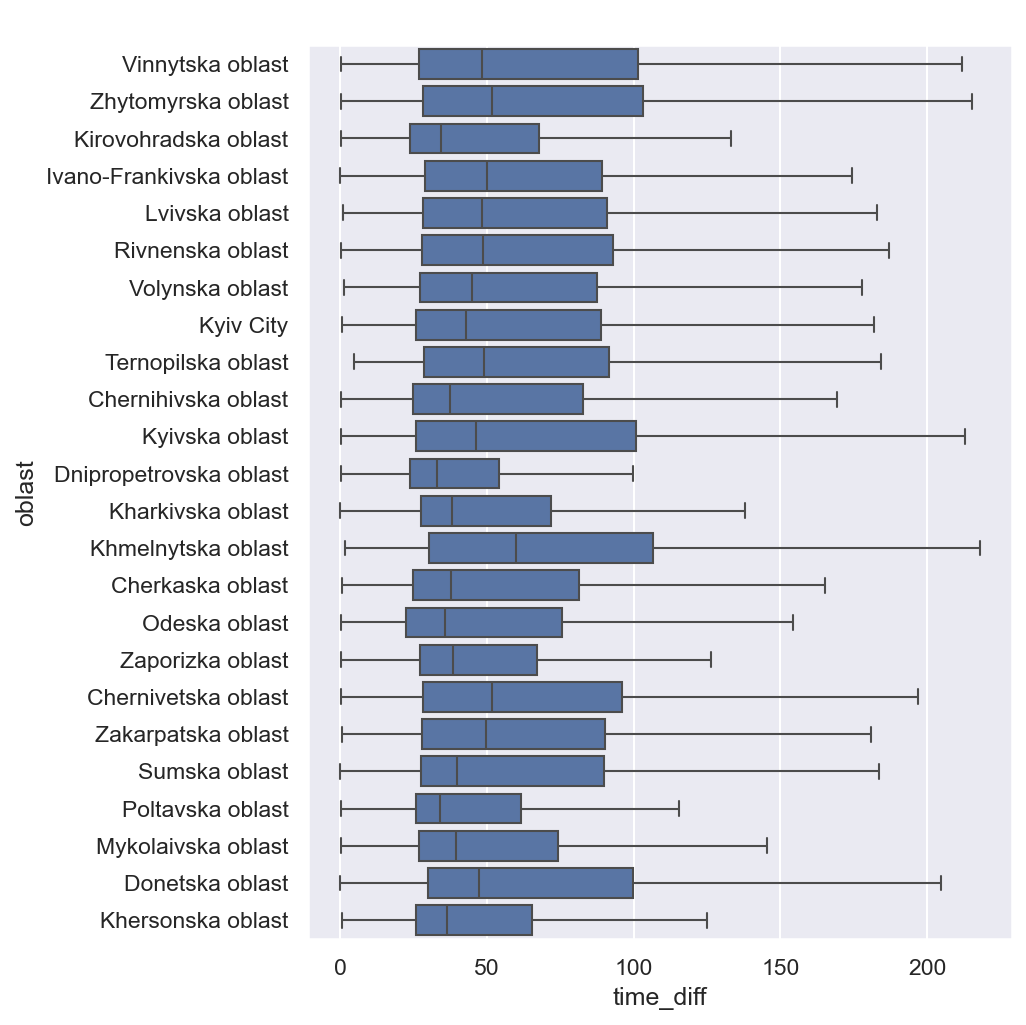}
 \caption{Boxplots for the duration of alerts by regions (minutes)}
 \label{boxplot_time_diff}
 \end{figure}

\begin{figure}[H]
\center
 \includegraphics[width=0.7\linewidth]{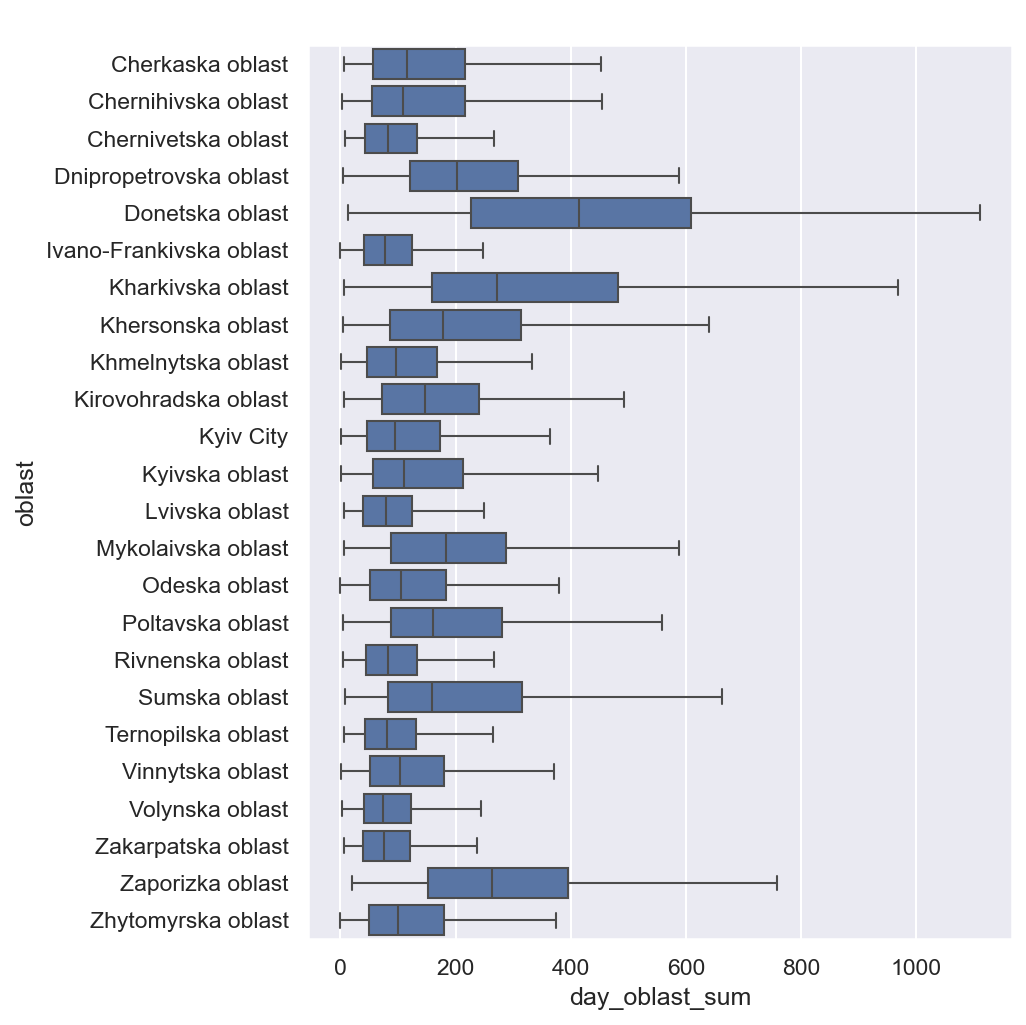}
 \caption{Boxplots for the total duration of air alerts per day by regions (minutes)}
 \label{boxplot_obalst_sum}
 \end{figure}

\begin{figure}[H]
\center
 \includegraphics[width=0.85\linewidth]{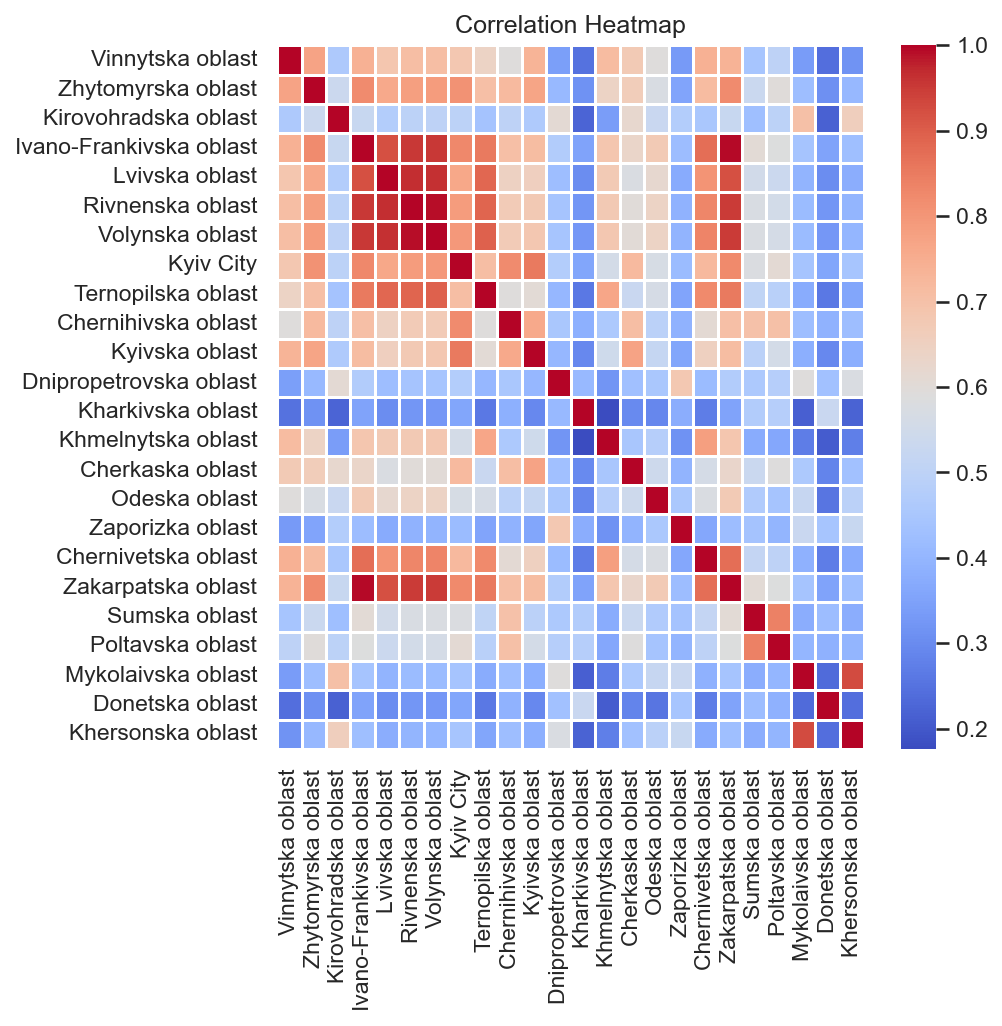}
 \caption{Correlation heatmap of time series of binary alert features for Ukrainian regions}
 \label{heatmap}
 \end{figure}

For our study, we took historical data from~\cite{airraiddatasets}. Here we present some results about exploratory data analysis. 
	Figure~\ref{time_distribution} shows the heatmap for the total duration of air alerts in Ukrainian regions (oblasts) in minutes. For Luhansk
	  region, the dataset contains only short time historical data.  
	  Figure~\ref{kharkiv_alert_conjc} shows a heatmap for regions' total duration of air alerts which occurred simultaneously with Kharkiv oblast. One can see that the alerts from neighbouring regions have higher correlation with air alerts in Kharkiv oblast if compared to more distant Western regions. 
	 Figure~\ref{median_time_diff} shows the daily median for alert durations by regions in minutes.
	Figure~\ref{boxplot_time_diff} shows the boxplots for the duration of alerts by regions in minutes. 
	Figure~\ref{boxplot_obalst_sum} shows the boxplots for the total duration of air alerts per day by regions in minutes.
	For further analysis, we created an \verb|alert_ts| dataframe with the time series index from Mar 25 2022 to Nov 6 2024 with the granularity of 1 minute. As binary feature names, we used the names of Ukrainian regions. In the case when an alert takes place at a certain time in a certain region, we set up value 1 for this feature, otherwise the value is 0.
	Figure~\ref{heatmap} shows the correlation heatmap of time series of binary alert features for Ukrainian regions using the  \verb|alert_ts| dataframe. One can see that alert time series of some regions has high level of correlation. It means there are geospatial patterns in air alerts. As a result, features from one region can have predictive potential for the air alert target variable for other regions. 

\FloatBarrier
 
\section{Training and validation of machine learning model}

\begin{figure}[H]
\center
 \includegraphics[width=1\linewidth]{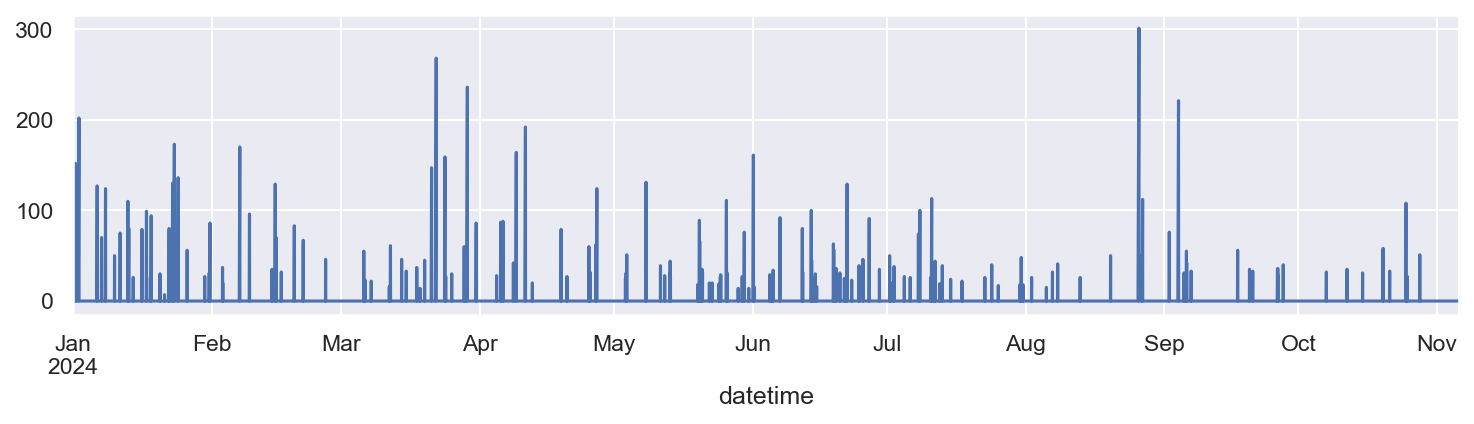}
 \caption{Alert time feature for Lvivska oblast (minutes)}
 \label{feat_Lvivska}
 \end{figure}

\begin{figure}[H]
\center
 \includegraphics[width=1\linewidth]{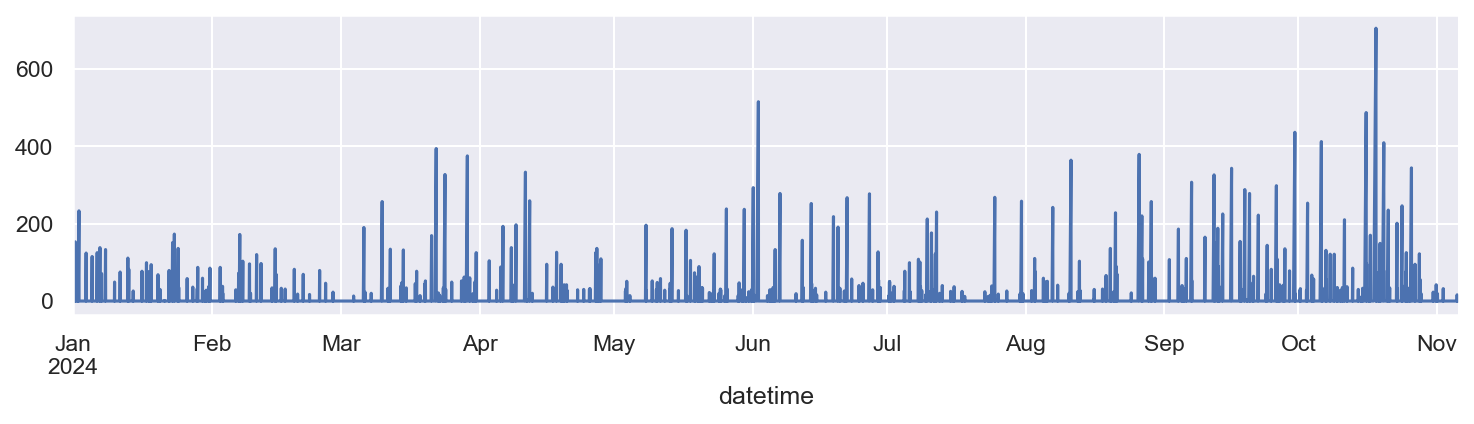}
 \caption{Alert time feature for Vinnytska oblast (minutes)}
 \label{feat_Vinnytska}
 \end{figure}

\begin{figure}[H]
\center
 \includegraphics[width=1\linewidth]{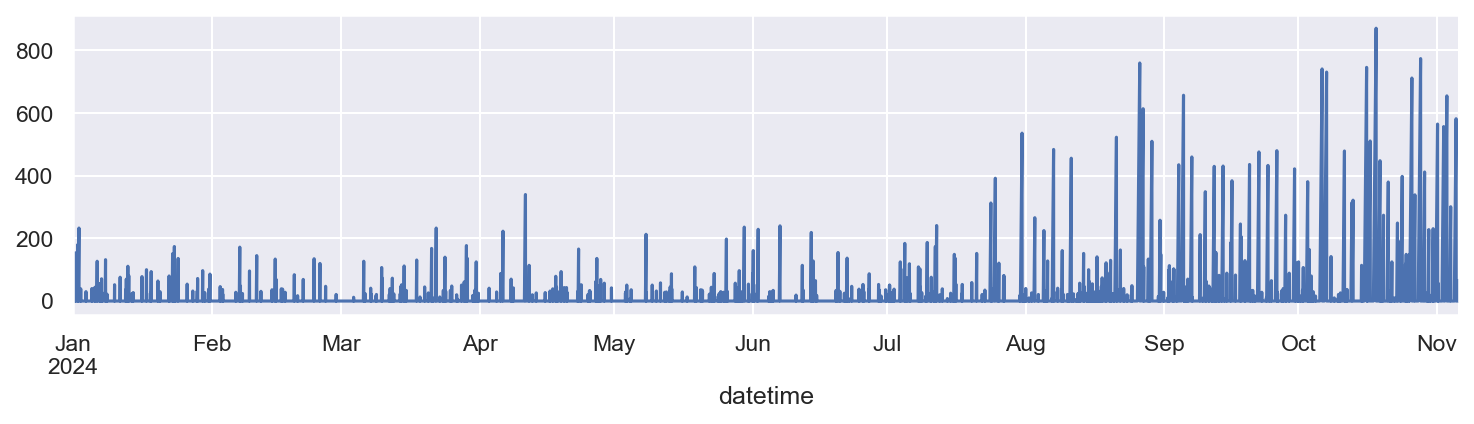}
 \caption{Alert time feature for Kyivska oblast (minutes)}
 \label{feat_Kyivska}
 \end{figure}

\begin{figure}[H]
\center
 \includegraphics[width=1\linewidth]{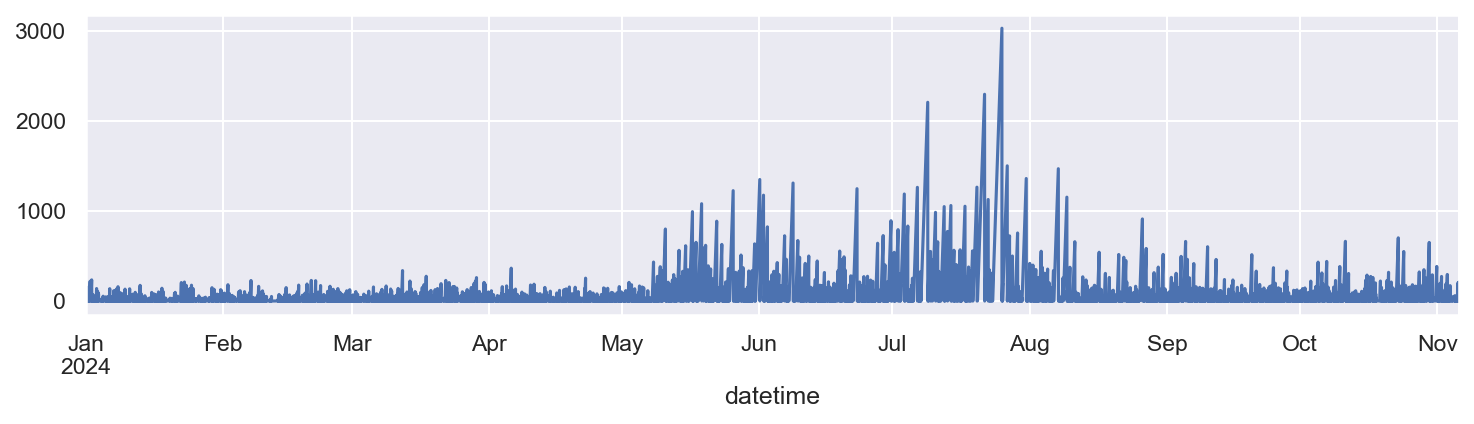}
 \caption{Alert time feature for Kharkivska oblast (minutes)}
 \label{feat_Kharkivska}
 \end{figure}

To build a predictive model, we can use binary features of the \verb|alert_ts| dataframe; as a target for the model, we can specify the binary variable that the alert is going to occur at some time period within next 5 minutes. So, value 1 means that the alert will take place within next 5 minutes. In the case when the alert is not occurring at the current moment, it means that it will start not later than 5 minutes. In the case when the alert is ongoing, it means that the alert will last for at least 5 minutes. If the target value is 0, it means that the alert is not going to start or will be over in 5 minutes.
	To build a predictive model, we can use lagged values for binary regions features. So, we can face with a large number of lagged values. We tried another approach based on the features with cumulative values for alert duration. When there is no air alert, the value for this feature is 0. When the air alert starts in a region, then this feature is equal to the current duration of the alert. When the alert in the region is over, then the feature will have the value  0.
For predictive analytics study, we chose the following regions: 
\textit{Lvivska oblast,  Vinnytska oblast, Kyivska oblast,  Kharkivska oblast}.
Figures~\ref{feat_Lvivska}--\ref{feat_Kharkivska} show  alert time features for these regions.
	
\FloatBarrier

\begin{figure}[H]
\center
 \includegraphics[width=0.75\linewidth]{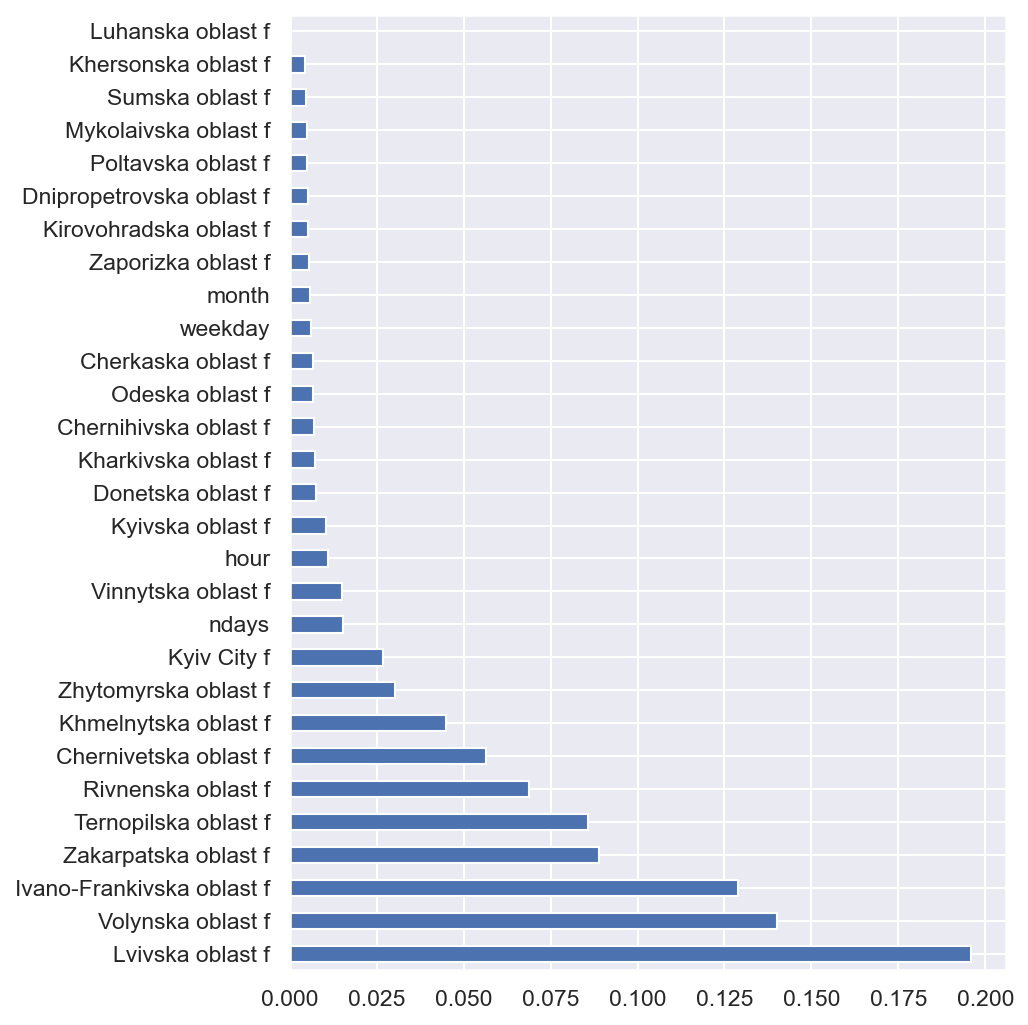}
 \caption{Features importance for Lvivska oblast (5 minute time horizon of target variable)}
 \label{i_Lvivska_5_imp}
 \end{figure}

\begin{figure}[H]
\center
 \includegraphics[width=0.5\linewidth]{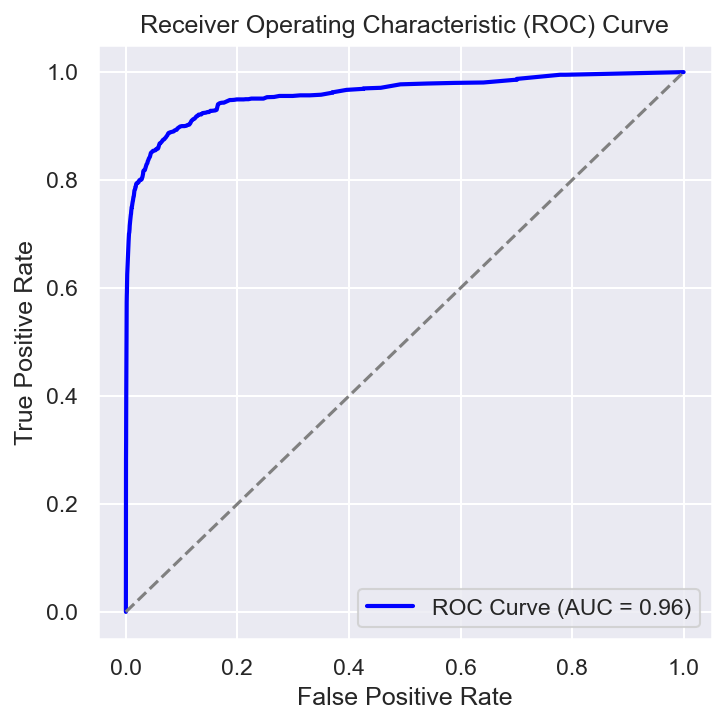}
 \caption{ROC curve for Lvivska  oblast (5 minute time horizon of target variable)}
 \label{i_Lvivska_5_roc_curve}
 \end{figure}

\begin{figure}[H]
\center
 \includegraphics[width=0.7\linewidth]{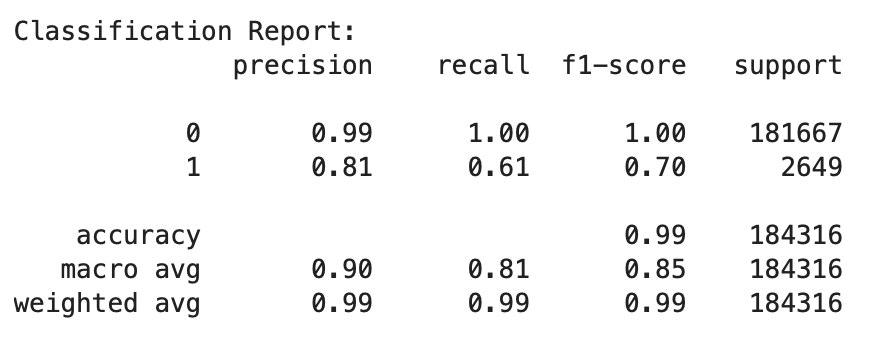}
 \caption{Accuracy scores for Lvivska oblast (5 minute time horizon of target variable)}
 \label{acc_Lvivska}
 \end{figure}

\begin{figure}[H]
\center
 \includegraphics[width=0.75\linewidth]{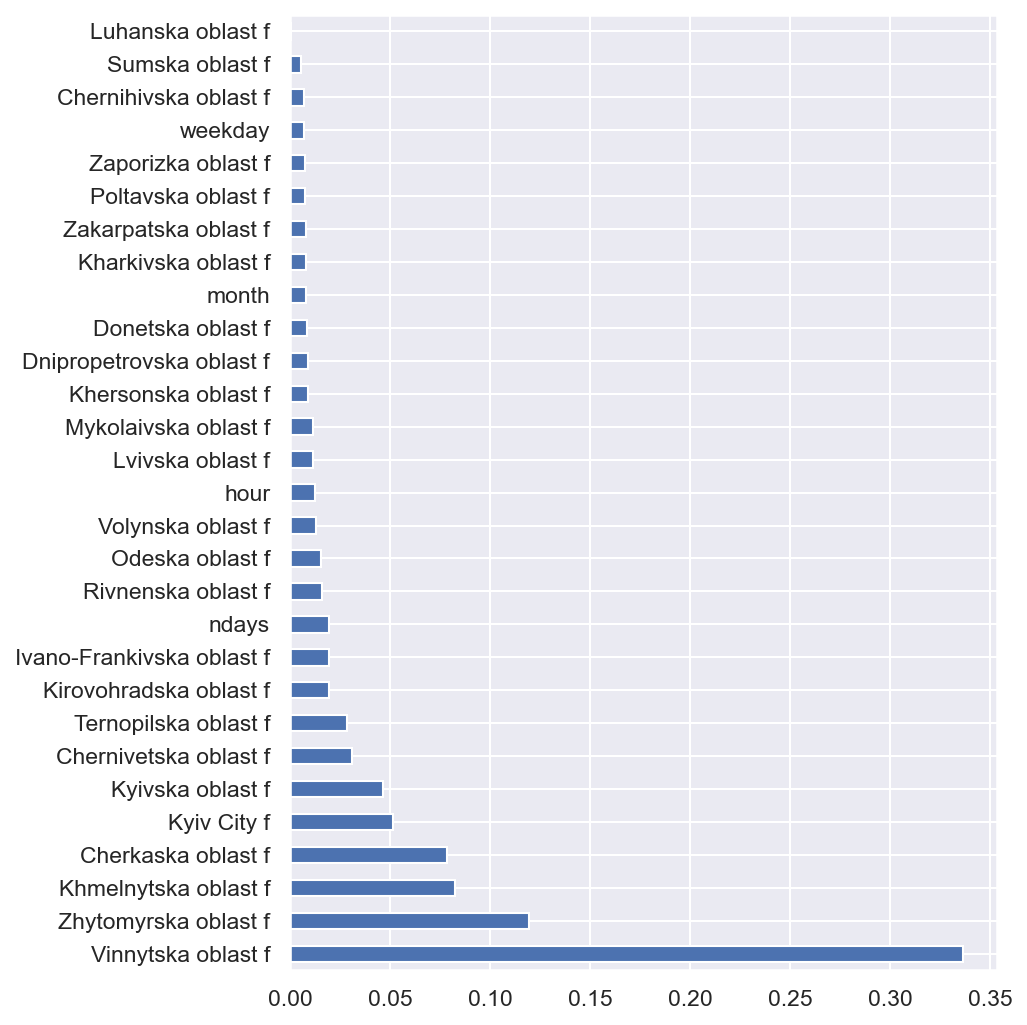}
 \caption{Features importance for Vinnytska oblast (5 minute time horizon of target variable)}
 \label{i_Vinnytska_5_imp}
 \end{figure}

\begin{figure}[H]
\center
 \includegraphics[width=0.5\linewidth]{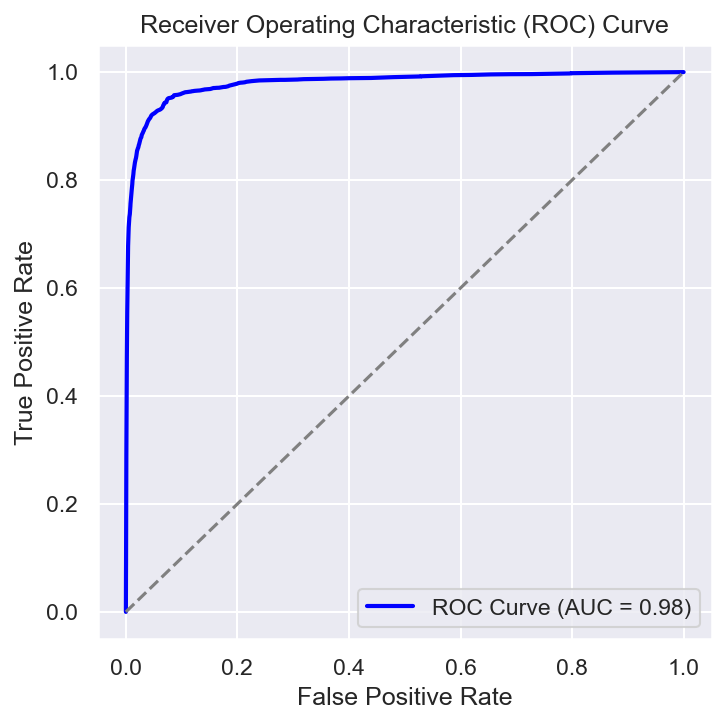}
 \caption{ROC curve for Vinnytska oblast (5 minute time horizon of target variable)}
 \label{i_Vinnytska_5_roc_curve}
 \end{figure}

\begin{figure}[H]
\center
 \includegraphics[width=0.7\linewidth]{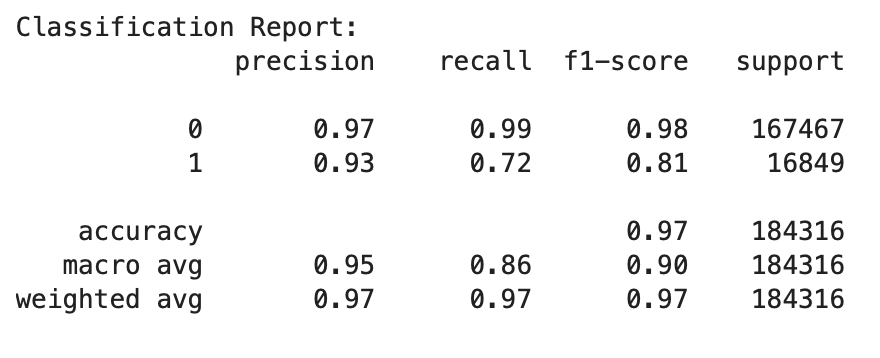}
 \caption{Accuracy scores for Vinnytska oblast (5 minute time horizon of target variable)}
 \label{acc_Vinnytska}
 \end{figure}

\begin{figure}[H]
\center
 \includegraphics[width=0.75\linewidth]{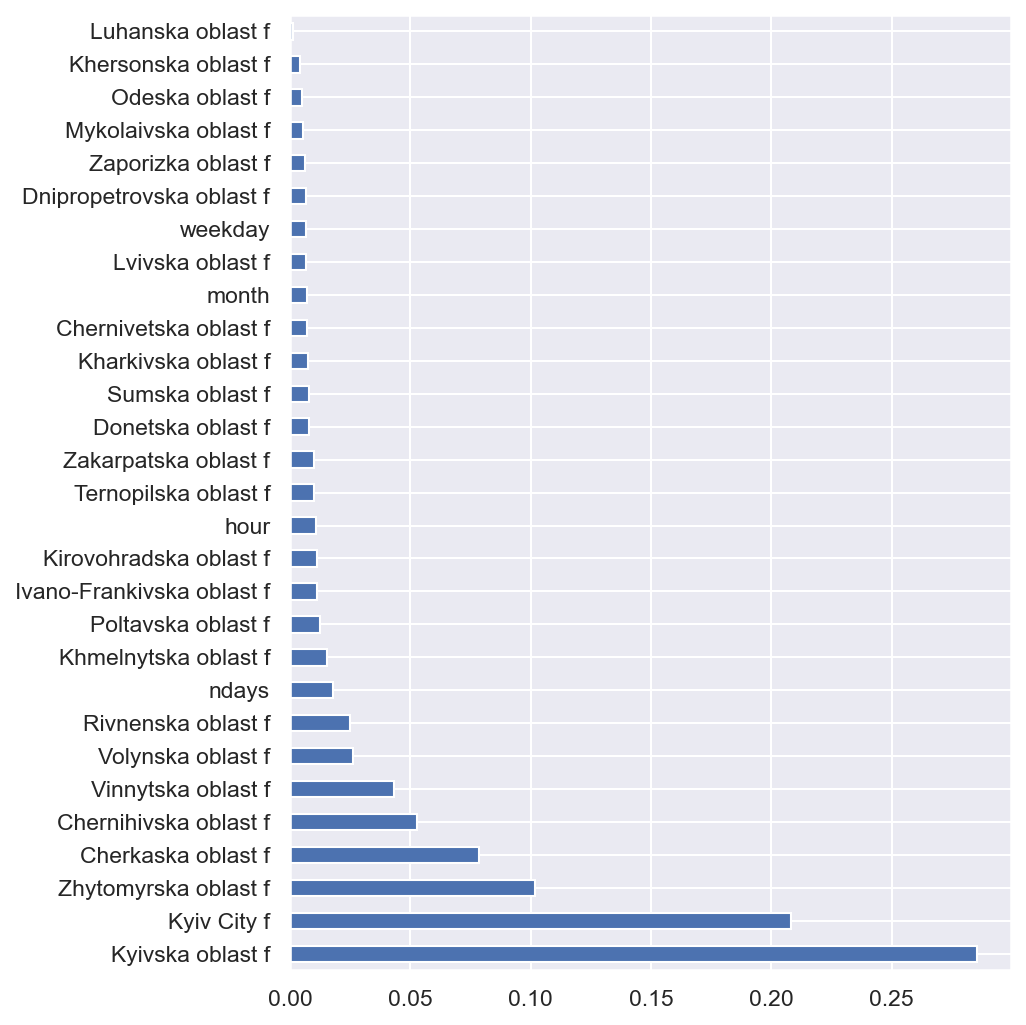}
 \caption{Features importance for Kyivska oblast (5 minute time horizon of target variable)}
 \label{i_Kyivska_5_imp}
 \end{figure}

\begin{figure}[H]
\center
 \includegraphics[width=0.5\linewidth]{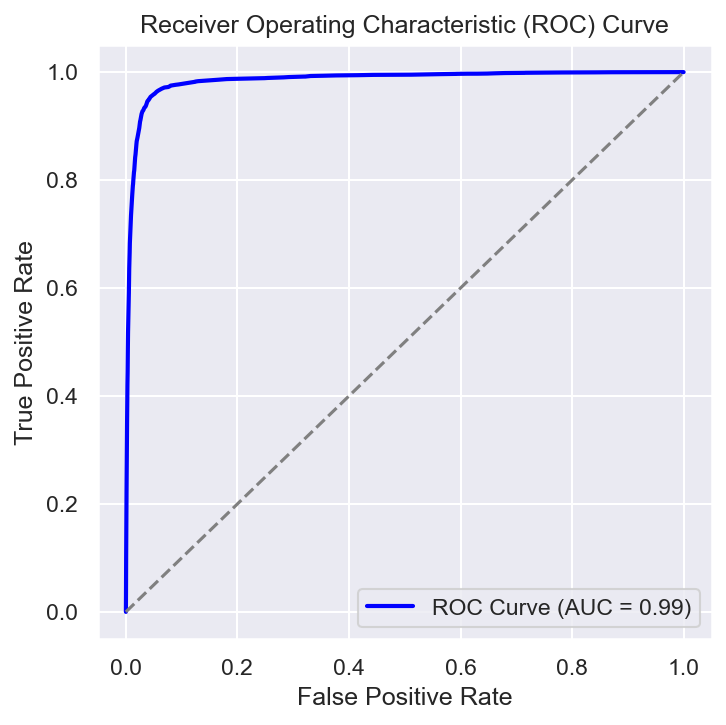}
 \caption{ROC curve for Kyivska oblast (5 minute time horizon of target variable)}
 \label{i_Kyivska_5_roc_curve}
 \end{figure}

\begin{figure}[H]
\center
 \includegraphics[width=0.7\linewidth]{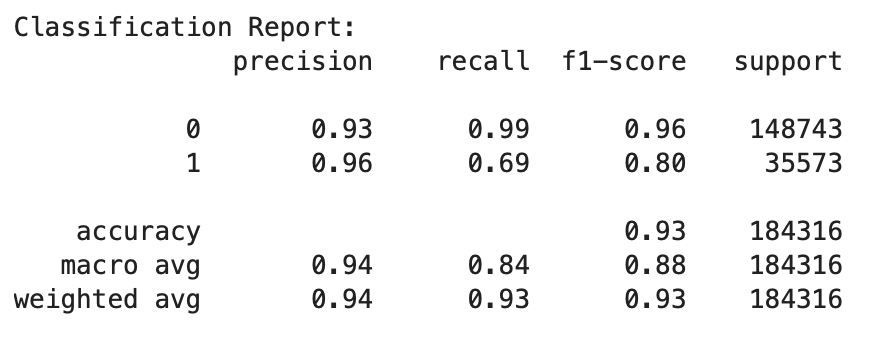}
 \caption{Accuracy scores for Kyivska oblast (5 minute time horizon of target variable)}
 \label{acc_Kyivska}
 \end{figure}

\begin{figure}[H]
\center
 \includegraphics[width=0.75\linewidth]{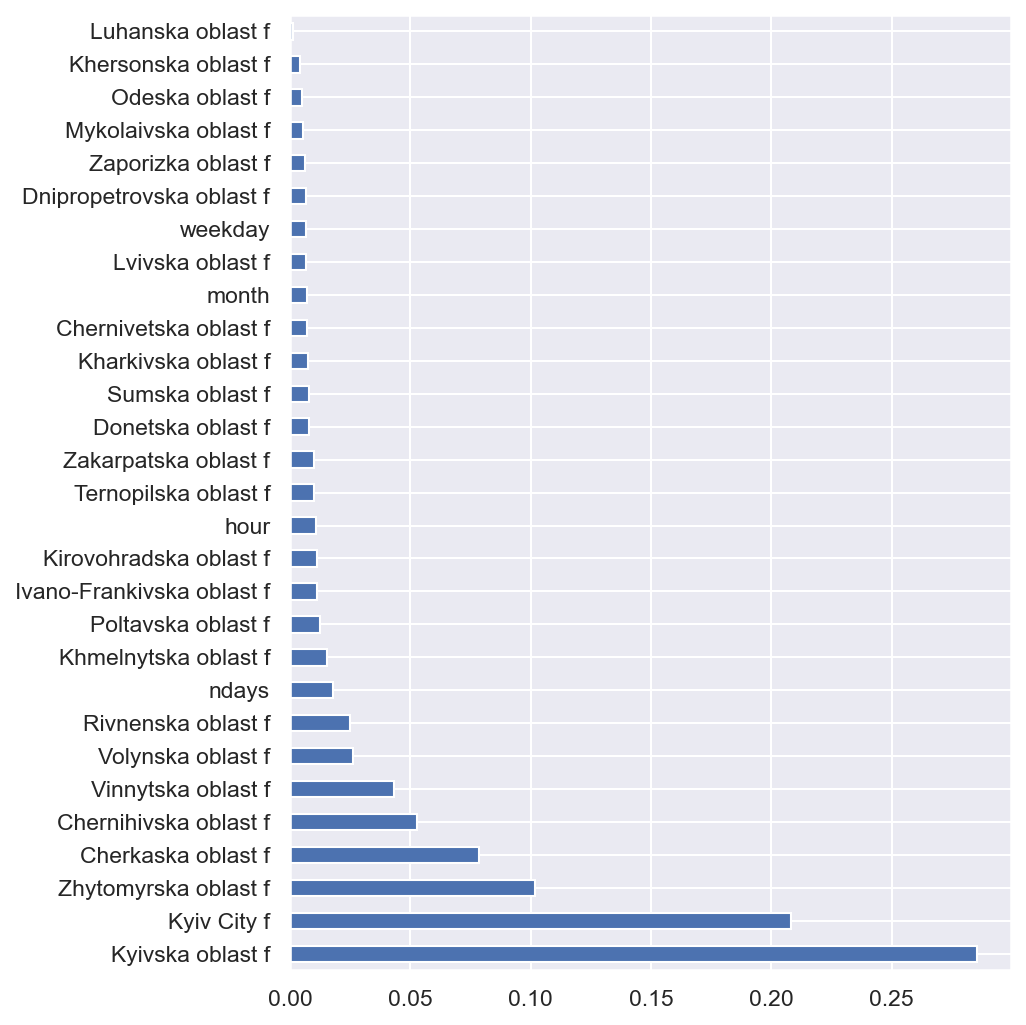}
 \caption{Features importance for Kharkivska oblast (5 minute time horizon of target variable)}
 \label{i_Kharkivska_5_imp}
 \end{figure}

\begin{figure}[H]
\center
 \includegraphics[width=0.5\linewidth]{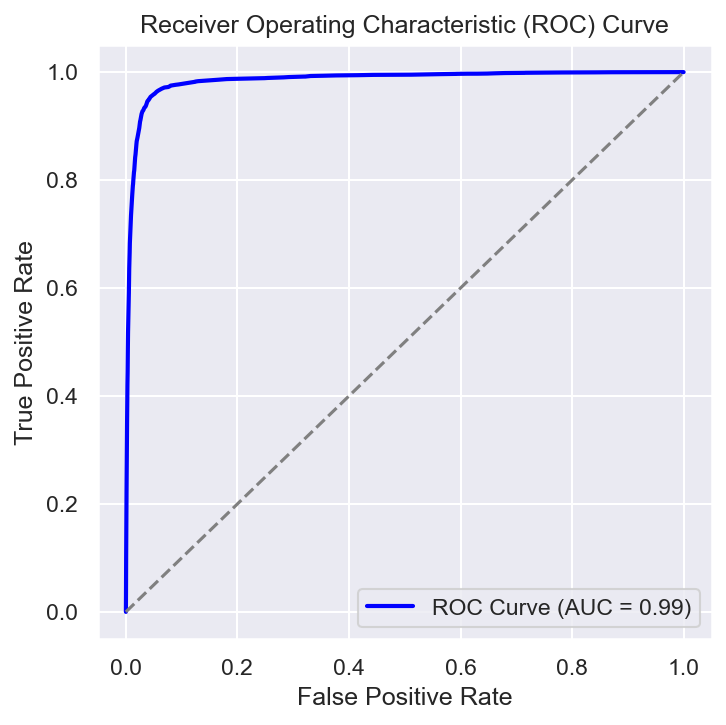}
 \caption{ROC curve for Kharkivska oblast (5 minute time horizon of target variable)}
 \label{i_Kharkivska_5_roc_curve}
 \end{figure}

\begin{figure}[H]
\center
 \includegraphics[width=0.7\linewidth]{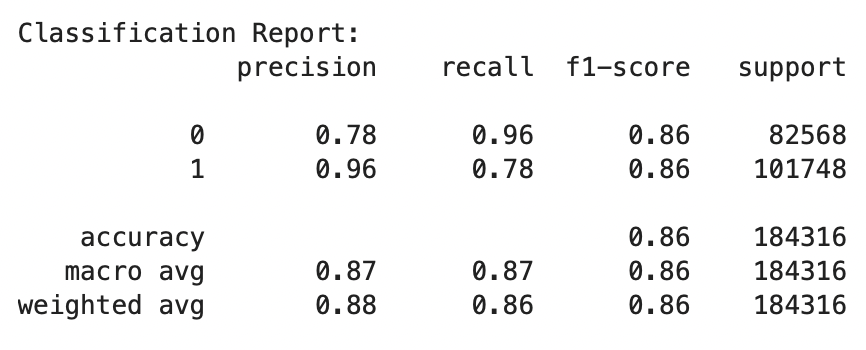}
 \caption{Accuracy scores for Kharkivska oblast (5 minute time horizon of target variable)}
 \label{acc_Kharkivska}
 \end{figure}

For each region under consideration, we have to build a separate model. For the classification predictive model, we used the Random Forest algorithm from \textit{scikit-learn} Python library. For training and validation, we created training and test datasets by splitting data by the time point 2024-07-01. The data before this time point belong to the train dataset, the data after this point belong to the test dataset for validation. Python function for training and validation of the machine learning model is in the Appendix. 
	For the training, we used 500 iterations. Figures~\ref{i_Lvivska_5_imp}--\ref{acc_Kharkivska} show  
	features importance, ROC curve and accuracy scores on the test dataset  for the regions under consideration in case of 5 minute time horizon of the target variable. Similar results for the case  of 15 minute time horizon of the target variable are shown in Figures~\ref{i_Lvivska_15_imp}--\ref{acc_Kharkivska_15} in Appendix.
	As the results show, the features of neighbouring regions play more important role than other features.
	We also used date and time features  like month, day of week, hour and time feature \verb|ndays| which means the number of days from the initial date in the dataset. The importance of this feature is caused by the fact that the patterns of air alerts are constantly changing and there are different alert patterns for different time periods.

\section{Conclusion}
In this study, we considered exploratory data analysis and approaches of predictive analytics for air alerts during Russian-Ukrainian war which started on Feb 24, 2022. The results show that alerts in some regions highly correlate with one another and have  patterns which make it possible to build a predictive model to predict that alert in a certain region will start within a certain time period. Obtained results show that features from neighboring regions play more important role for the target variable of a specified region. Seasonality features such as hour, day of week and month, as well as  time features equal to number of days from the initial day of dataset are also of high importance. We can observe that air alert patterns change with time.  Alert time series of some regions have high level of correlation. It means that there are geospatial patterns in air alerts. As a result, features from one region can have a predictive potential for the air alert target variable in other regions. 

\bibliographystyle{unsrt}
\bibliography{article.bib}
\FloatBarrier
\appendix
\section{Appendix}
\subsection{Forecasting alerts with time horizon 15 minutes}
\begin{figure}[H]
\center
 \includegraphics[width=0.7\linewidth]{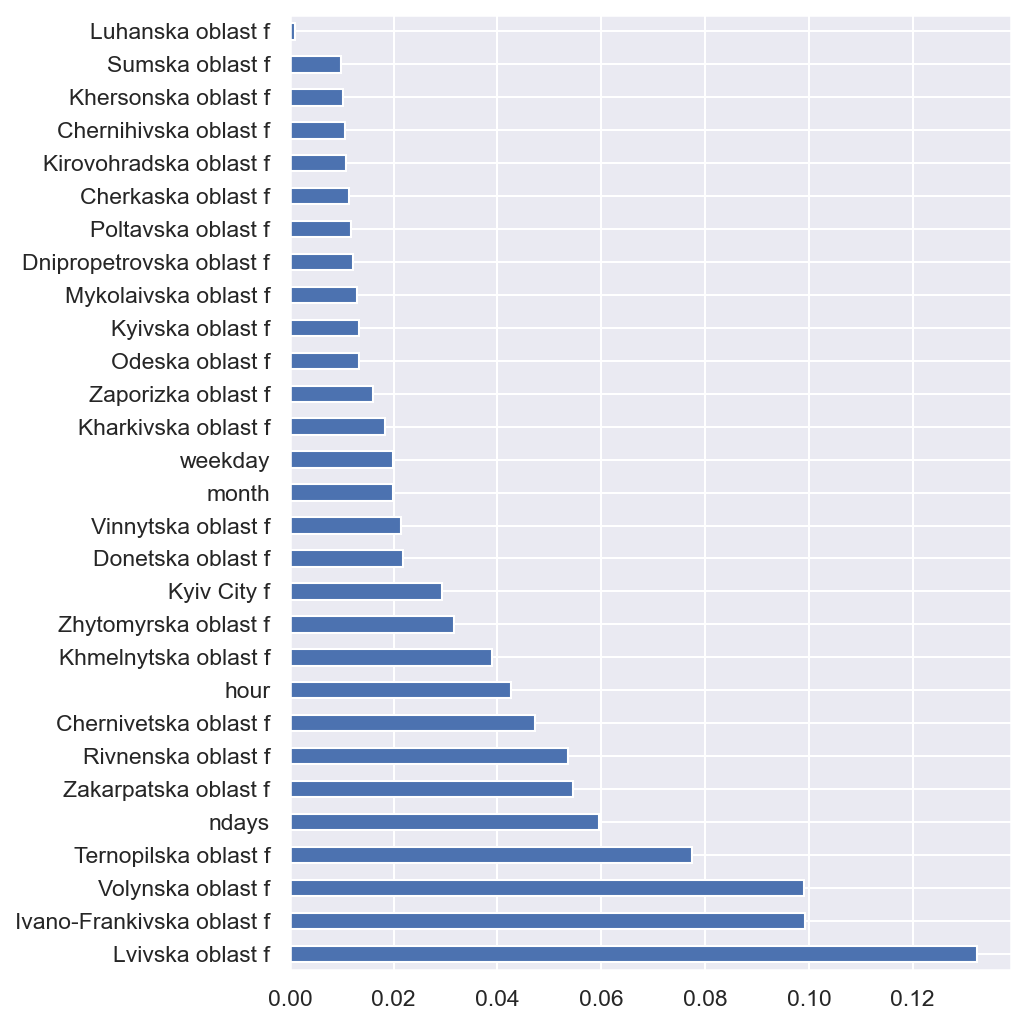}
 \caption{Features importance for Lvivska oblast (15 minute time horizon of the target variable)}
 \label{i_Lvivska_15_imp}
 \end{figure}

\begin{figure}[H]
\center
 \includegraphics[width=0.5\linewidth]{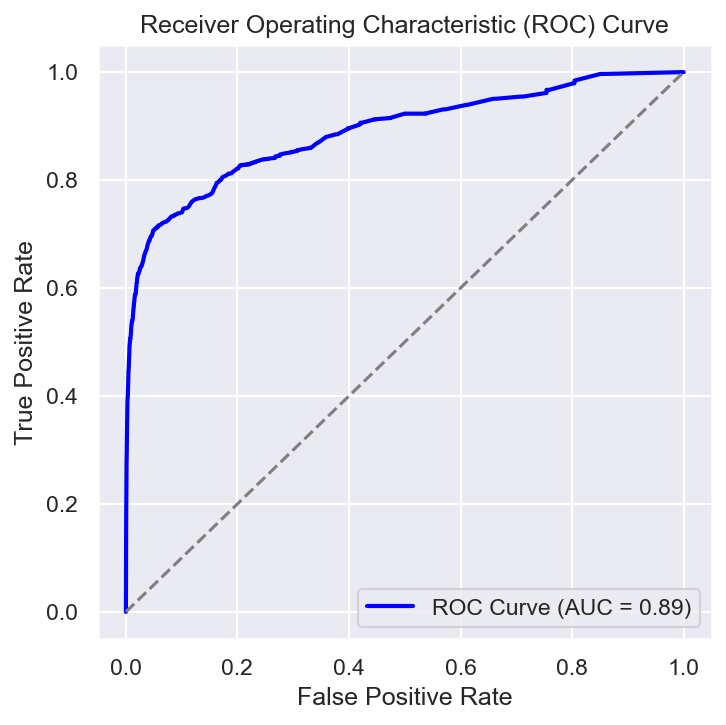}
 \caption{ROC curve for Lvivska oblast (15 minute time horizon of the target variable)}
 \label{i_Lvivska_15_roc_curve}
 \end{figure}

\begin{figure}[H]
\center
 \includegraphics[width=0.7\linewidth]{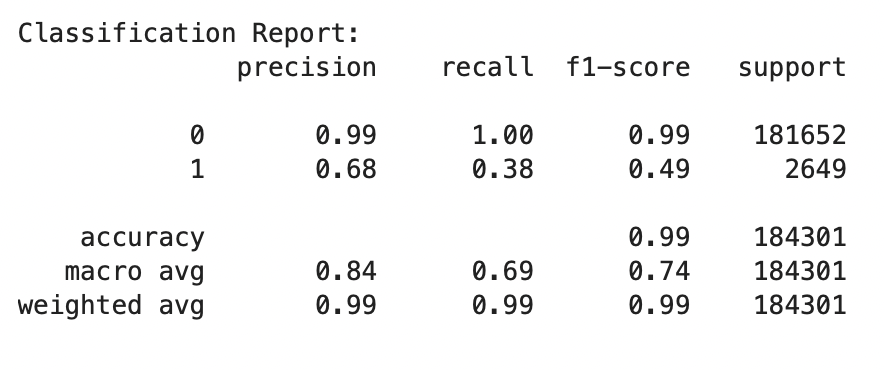}
 \caption{Accuracy scores for Lvivska oblas (15 minute time horizon of the target variable)}
 \label{acc_Lvivska_15}
 \end{figure}

\begin{figure}[H]
\center
 \includegraphics[width=0.7\linewidth]{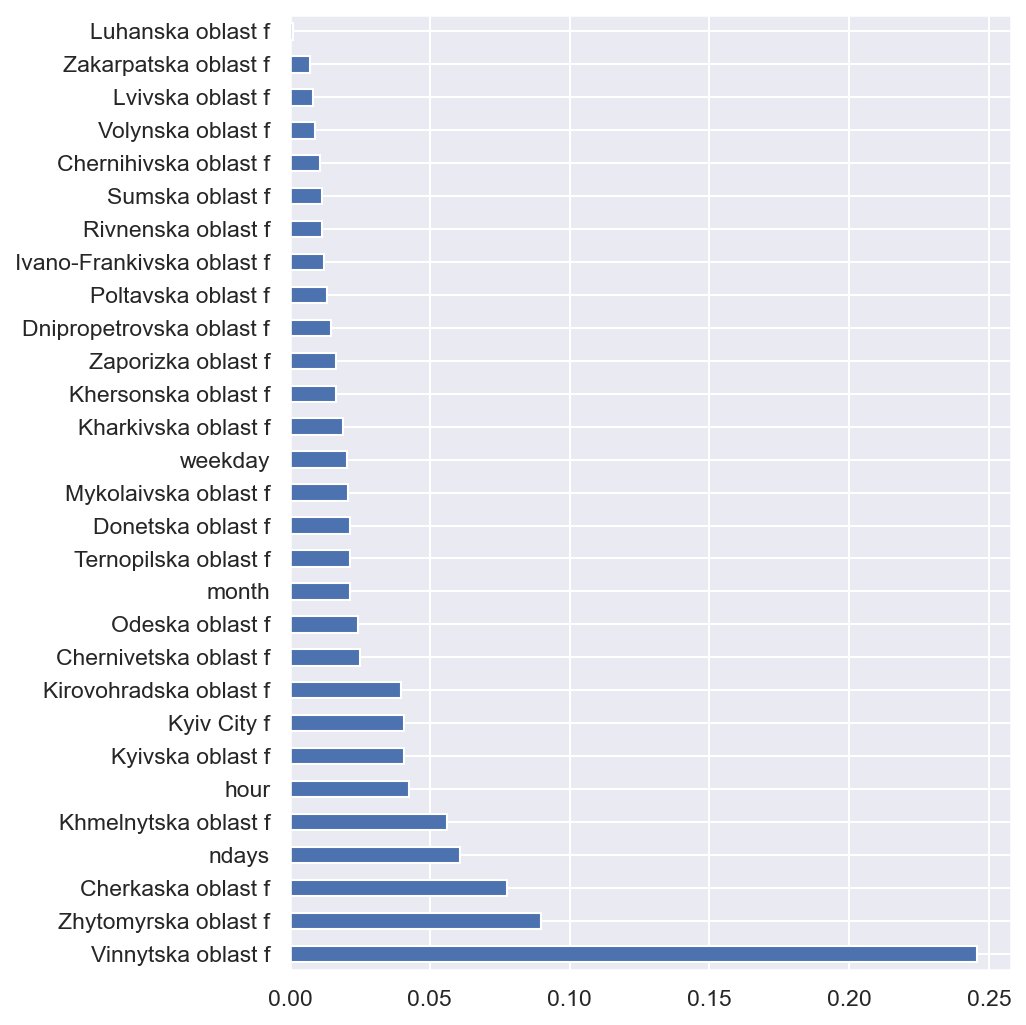}
 \caption{Features importance for Vinnytska oblast (15 minute time horizon of the target variable)}
 \label{i_Vinnytska_15_imp}
 \end{figure}

\begin{figure}[H]
\center
 \includegraphics[width=0.5\linewidth]{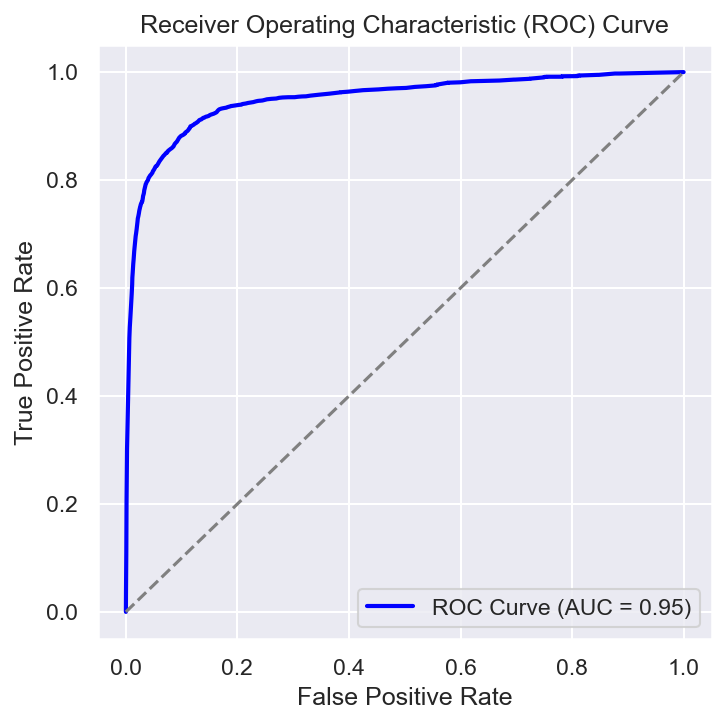}
 \caption{ROC curve for Vinnytska oblast (15 minute time horizon of the target variable)}
 \label{i_Vinnytska_15_roc_curve}
 \end{figure}

\begin{figure}[H]
\center
 \includegraphics[width=0.7\linewidth]{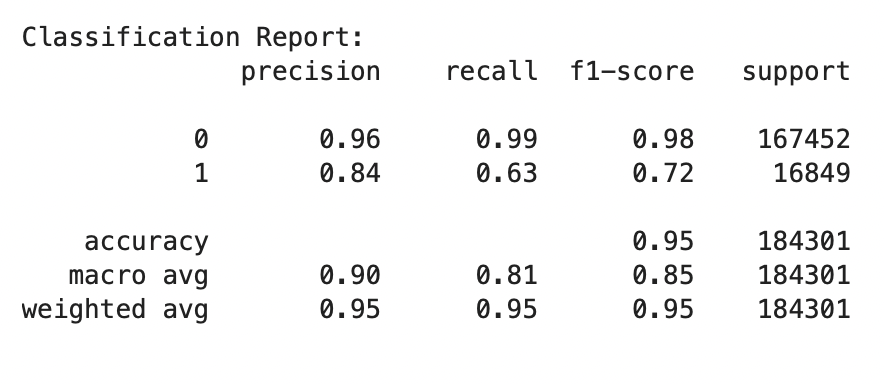}
 \caption{Accuracy scores for Vinnytska oblast (15 minute time horizon of the target variable)}
 \label{acc_Vinnytska_15}
 \end{figure}

\begin{figure}[H]
\center
 \includegraphics[width=0.7\linewidth]{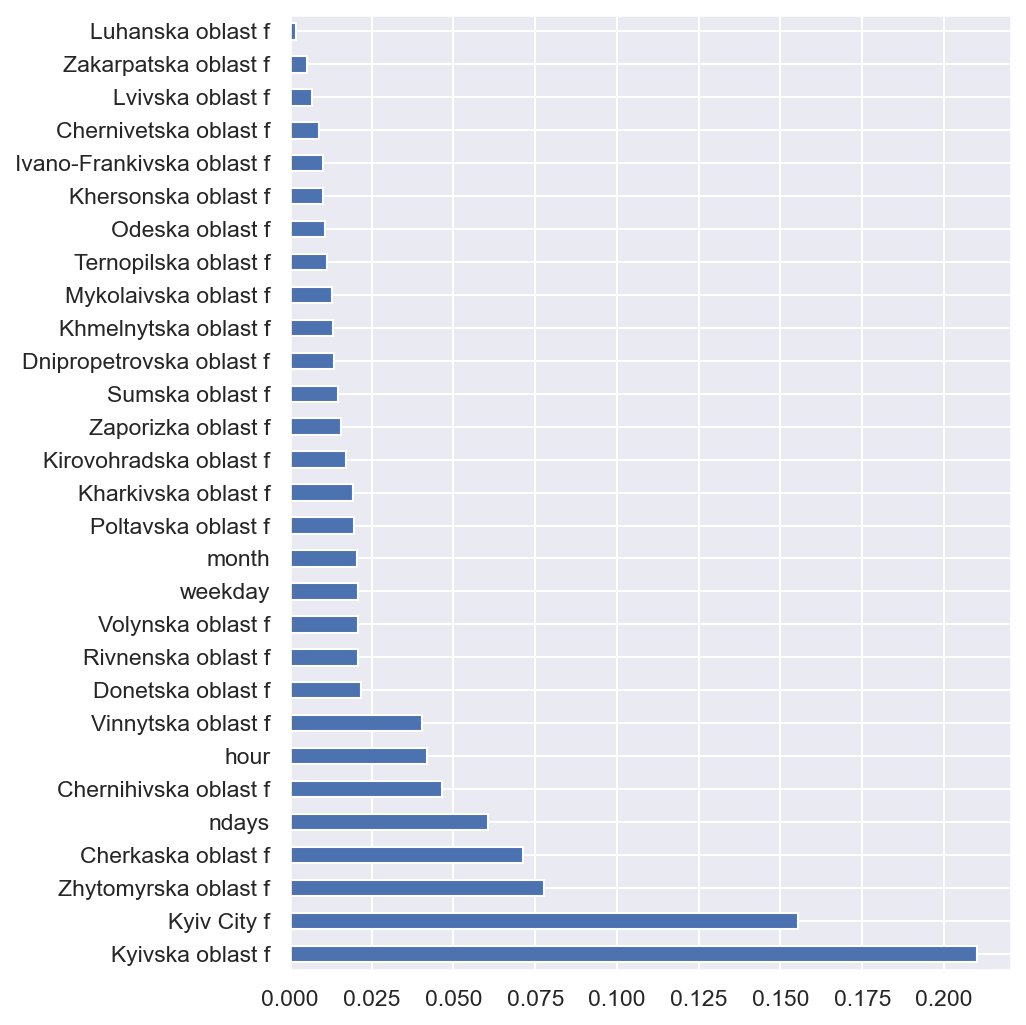}
 \caption{Features importance for Kyivska oblast (15 minute time horizon of the target variable)}
 \label{i_Kyivska_15_imp}
 \end{figure}

\begin{figure}[H]
\center
 \includegraphics[width=0.5\linewidth]{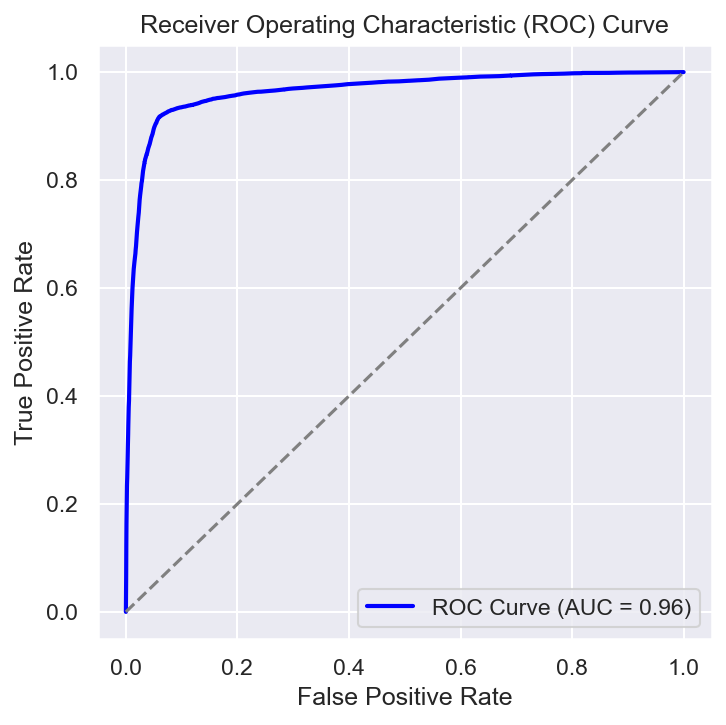}
 \caption{Features importance for Kyivska oblast (15 minute time horizon of the target variable)}
 \label{i_Kyivska_15_roc_curve}
 \end{figure}

\begin{figure}[H]
\center
 \includegraphics[width=0.7\linewidth]{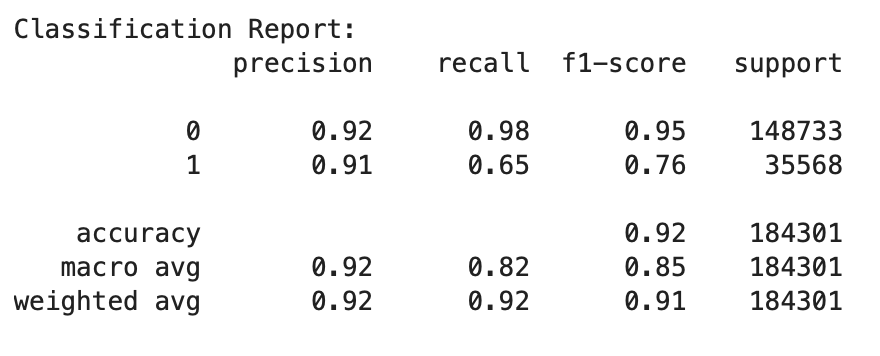}
 \caption{Accuracy scores for Kyivska oblast (15 minute time horizon of the target variable)}
 \label{acc_Kyivska_15}
 \end{figure}

\begin{figure}[H]
\center
 \includegraphics[width=0.7\linewidth]{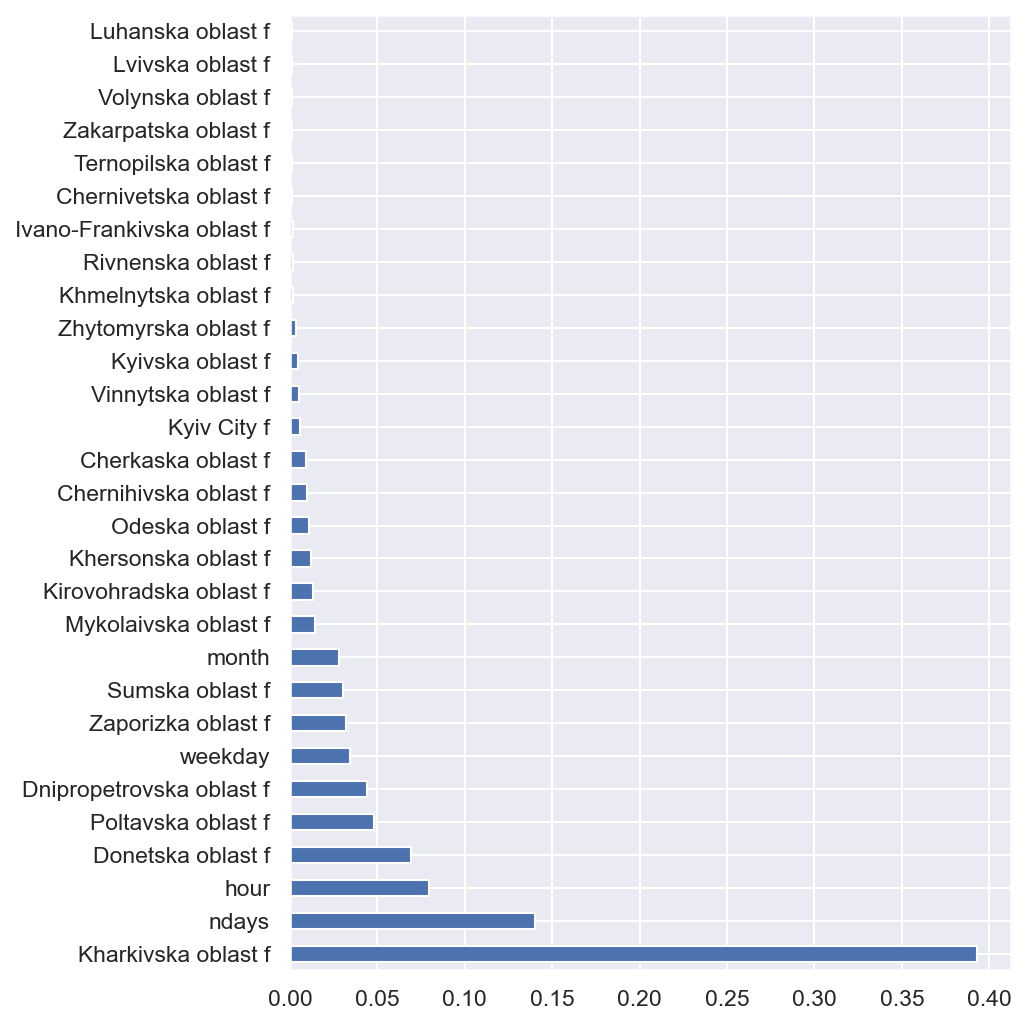}
 \caption{Features importance for Kharkivska oblast (15 minute time horizon of the target variable)}
 \label{i_Kharkivska_15_imp}
 \end{figure}

\begin{figure}[H]
\center
 \includegraphics[width=0.5\linewidth]{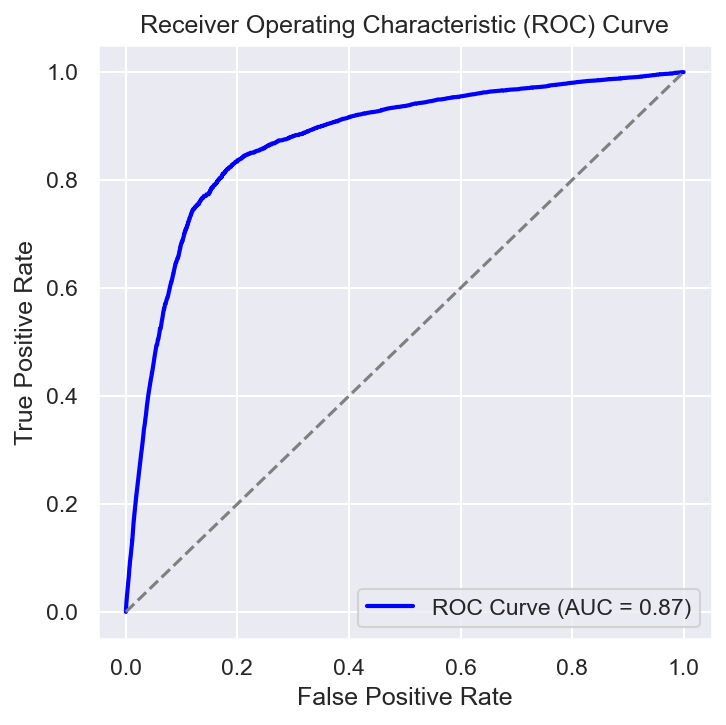}
 \caption{ROC curve for Kharkivska oblast (15 minute time horizon of the target variable)}
 \label{i_Kharkivska_15_roc_curve}
 \end{figure}

\begin{figure}[H]
\center
 \includegraphics[width=0.7\linewidth]{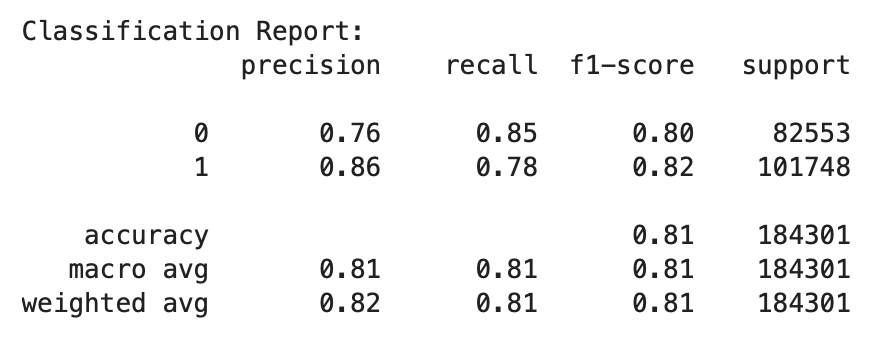}
 \caption{Accuracy scores for Kharkivska oblast (15 minute time horizon of the target variable)}
 \label{acc_Kharkivska_15}
 \end{figure}

Figures~\ref{i_Lvivska_15_imp}--\ref{acc_Kharkivska_15} show 
	features importance, ROC curve and accuracy scores on the test dataset for the regions under investigations in case of 15 minute time horizon of target variable.

\subsection{Python function for training and validation of machine learning model}
\begin{adjustwidth}{-1cm}{ -2cm}
\lstset{
  basicstyle  = \fontfamily{pcr}\fontsize{9pt}{10pt}\selectfont,
  language    = Python,
  breaklines=true
}
\begin{lstlisting}[language = Python]
def train_validate_model(alert_ts, features, target, train_test_split_date,
                         n_estimators, filename):
    
    """
    Training and validation machine learning model for prediction of alert duration 
    """
    
    alert_ts[target] = alert_ts[target].astype(int)
    alert_ts_train = alert_ts[alert_ts['datetime']<train_test_split_date]
    alert_ts_test = alert_ts[alert_ts['datetime']>=train_test_split_date]
    X_train, X_test  = alert_ts_train[features],alert_ts_test[features]
    y_train, y_test = alert_ts_train[target],alert_ts_test[target]
    cl = RandomForestClassifier(n_estimators=n_estimators, random_state=15)
    cl.fit(X_train, y_train)
    y_pred = cl.predict(X_test)
    
    accuracy = accuracy_score(y_test, y_pred)
    print(f"Accuracy: {accuracy:.2f}")
    print("\nClassification Report:")
    print(classification_report(y_test, y_pred))
    
    print("\nConfusion Matrix:")
    print(confusion_matrix(y_test, y_pred))
    
    y_prob = cl.predict_proba(X_test)[:, 1]
    
    feature_importances = pd.Series(cl.feature_importances_, index=features)
    feature_importances.sort_values(ascending=False, inplace=True)
    feature_importances.plot(kind='barh',figsize=(7,7))
    plt.tight_layout()
    plt.savefig(filename+'_imp.png', dpi=150, bbox_inches='tight')
    plt.show()
    
    fpr, tpr, _ = roc_curve(y_test, y_prob)
    roc_auc = auc(fpr, tpr)
    
    # Plot the ROC curve
    plt.figure(figsize=(5, 5))
    plt.plot(fpr, tpr, color="blue", lw=2, label=f"ROC Curve (AUC = {roc_auc:.2f})")
    plt.plot([0, 1], [0, 1], color="gray", linestyle="--")  
    plt.xlabel("False Positive Rate")
    plt.ylabel("True Positive Rate")
    plt.title("Receiver Operating Characteristic (ROC) Curve")
    plt.legend(loc="lower right")
    plt.tight_layout()
    plt.savefig(filename+'_roc_curve.png', dpi=150, bbox_inches='tight')
    plt.show()
    return(alert_ts_train, alert_ts_test, y_pred, y_prob, cl)
\end{lstlisting}
\end{adjustwidth}

\end{document}